\documentclass{article}
\usepackage{ijcai24}

\usepackage[table,xcdraw]{xcolor}
\usepackage{tcolorbox}
\usepackage{times}
\usepackage{soul}
\usepackage{url}
\usepackage[hidelinks]{hyperref}
\usepackage[utf8]{inputenc}
\usepackage[small]{caption}
\usepackage{graphicx}
\usepackage{amsmath}
\usepackage{amsthm}
\usepackage{amsfonts}
\usepackage{booktabs}
\usepackage{algorithm}
\usepackage{algorithmic}
\usepackage[switch]{lineno}
\usepackage{enumitem}
\usepackage{multirow}
\usepackage{bm}

\usepackage{listings} 
\definecolor{codeblue}{rgb}{0.28,0.24,0.55}
\definecolor{codeorange}{rgb}{0.78,0.41,0.08}

\lstdefinestyle{mystyle}{
    basicstyle=\bfseries\footnotesize,
    commentstyle=\color{codeblue},
    keywordstyle=\color{codeorange},
    breakatwhitespace=false,         
    breaklines=true,                 
    captionpos=b,                    
    keepspaces=true,                 
    numbers=none,                    
    numbersep=5pt,                  
    showspaces=false,                
    showstringspaces=false,
    showtabs=false,                  
    tabsize=2,
    frame=tb, 
    framerule=0.4pt, 
    framesep=3pt, 
    lineskip=3pt
}

\title{Gradformer: Graph Transformer with Exponential Decay}

\author{
Chuang Liu$^1$\thanks{Equal Contribution}\and
Zelin Yao$^{1*}$\and
Yibing Zhan$^2$\and
Xueqi Ma$^3$\and
Shirui Pan$^4$\and
Wenbin Hu$^1$\footnote{Corresponding Author} 
\affiliations
$^1$School of Computer Science, Wuhan University, Wuhan, China\\
$^2$JD Explore Academy, JD.com, China \\
$^3$School of Computing and Information Systems, The University of Melbourne, Melbourne, Australia \\
$^4$School of Information and Communication Technology, Griffith University, Brisbane, Australia \\
\emails
\{chuangliu, zelinyao, hwb\}@whu.edu.cn,
zhanyibing@jd.com,
xueqim@student.unimelb.edu.au,
s.pan@griffith.edu.au
}

\begin{document}

\maketitle

\begin{abstract}
Graph Transformers (GTs) have demonstrated their advantages across a wide range of tasks. However, the self-attention mechanism in GTs overlooks the graph's inductive biases, particularly biases related to structure, which are crucial for the graph tasks. Although some methods utilize positional encoding and attention bias to model inductive biases, their effectiveness is still suboptimal analytically. Therefore, this paper presents Gradformer, a method innovatively integrating GT with the intrinsic inductive bias by applying an exponential decay mask to the attention matrix. Specifically, the values in the decay mask matrix diminish exponentially, correlating with the decreasing node proximities within the graph structure. This design enables Gradformer to retain its ability to capture information from distant nodes while focusing on the graph's local details. Furthermore, Gradformer introduces a learnable constraint into the decay mask, allowing different attention heads to learn distinct decay masks. Such an design diversifies the attention heads, enabling a more effective assimilation of diverse structural information within the graph. Extensive experiments on various benchmarks demonstrate that Gradformer consistently outperforms the Graph Neural Network and GT baseline models in various graph classification and regression tasks. Additionally, Gradformer has proven to be an effective method for training deep GT models, maintaining or even enhancing accuracy compared to shallow models as the network deepens, in contrast to the significant accuracy drop observed in other GT models.Codes are available at \url{https://github.com/LiuChuang0059/Gradformer}.
\end{abstract}







\section{Introduction}
\label{sec:introduction}


\begin{figure}[!t] 
\begin{center}
\includegraphics[width=1.0\linewidth]{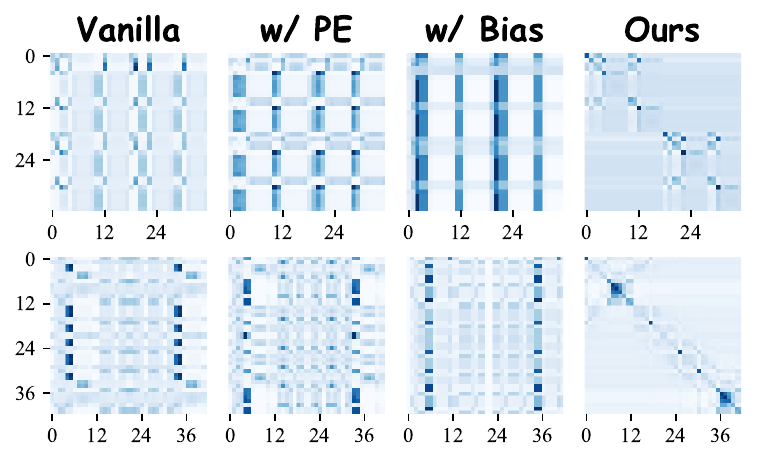}
\end{center}
\caption{Visualization of attention patterns in different GT models with two graphs from the OGBG-HIV dataset. From left to right: vanilla GT, GT with position encoding (\textit{w/ PE}), GT with attention bias (\textit{w/ Bias}), and GT with our proposed decay mask (\textit{ours}).}
\label{fig:attn_vis}
\end{figure}

Graph Transformers (GTs)~\cite{graph-transformer} have shown remarkable success in achieving state-of-the-art performance across various applications. Unlike the local message-passing in graph neural networks (GNNs), GTs can capture long-range information from distant nodes. Specifically, the self-attention mechanism in GTs allows each node to directly attend to other nodes in a graph, enabling information aggregation from arbitrary nodes. 

The self-attention in GTs offers considerable flexibility and the capacity to aggregate information both globally and adaptively. However, this mechanism significantly neglects intrinsic inductive biases in graphs, which poses challenges in capturing the essential graph structural information. Without considering structural relationships, the indiscriminate attention to all nodes in a graph will render the self-attention mechanism's inadequate focus on key information and the aggregation of redundant information. This comes with the generation of meaningless attention scores and, ultimately, suboptimal learning outcomes. Furthermore, the above issue is particularly evident in scenarios with limited data.

\begin{figure*}[!t] 
\begin{center}
\includegraphics[width=0.95\linewidth]{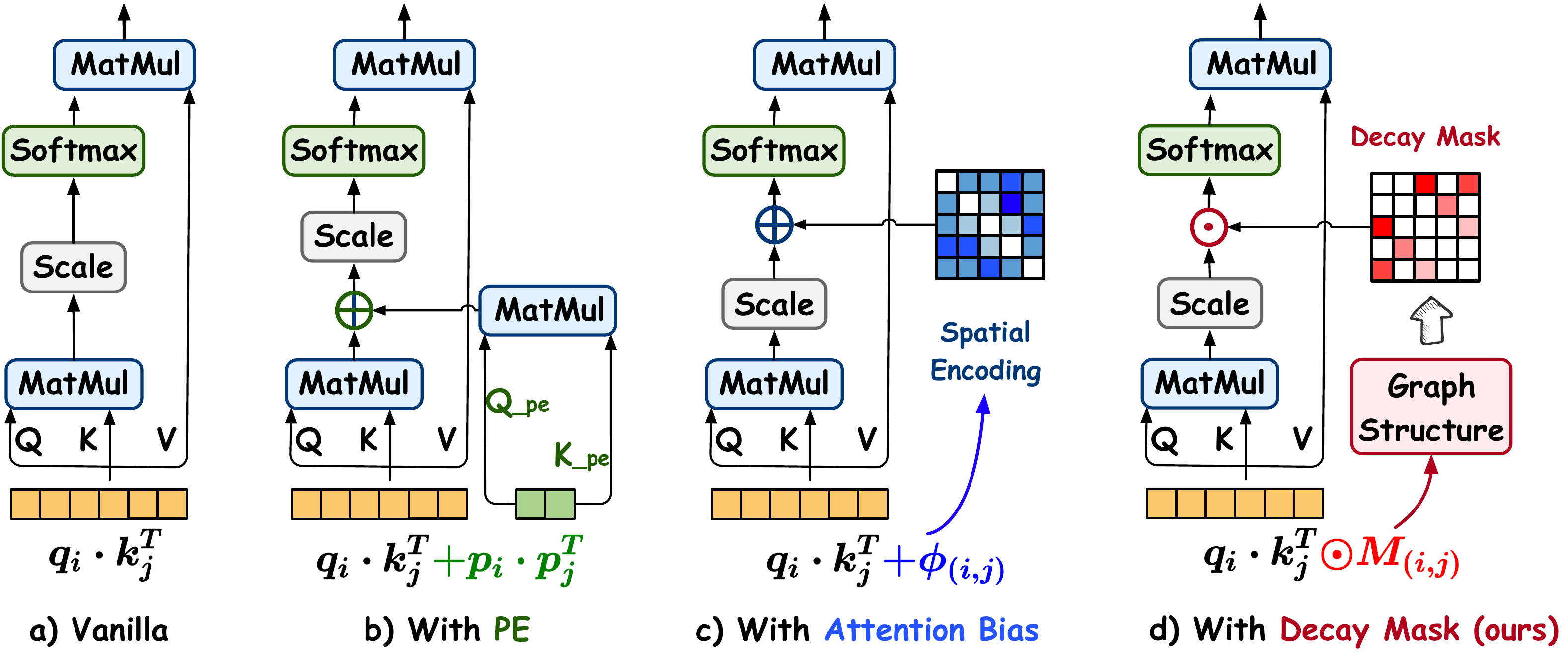}
\end{center}
\caption{The overview of the Gradformer framework and its comparison with existing methods. \textbf{a) Vanilla:} Vanilla self-attention mechanism serves as the baseline. \textbf{b) With PE:} In several works \protect\cite{graph-transformer}, the PE vector  (\textit{i.e.}, $\mathbf{p}_i$) is concatenated with the input node features, which can be interpreted as introducing a bias in the attention score. \textbf{c) With Attention Bias:} Some methods \protect\cite{graphormer-v1} incorporate an attention bias (\textit{i.e.}, $\phi(i,j)$) into the attention score calculation. This bias often derives from spatial information, such as the shortest path. \textbf{d) Ours:} Our method introduces an exponential decay mask that is multiplied with the attention scores. This mask is derived from the structural information of the graph. Moreover, different attention heads utilize distinct masks, made possible through learnable parameters. A comprehensive explanation of the symbols is provided in Section~\ref{sec:method}.}
\label{fig:model}
\end{figure*}

Numerous studies have focused on integrating graph inductive biases into self-attention learning mechanisms. These studies are broadly divided into two main approaches: 1) Injected position encoding: Studies such as GT~\cite{graph-transformer} and SAN~\cite{san} suggest using laplacian eigenfunctions as positional encodings (PEs) to contain the structural characteristics of graphs. 2) Attention bias:  Graphormer~\cite{graphormer-v1} proposes a direct addition of structural encodings into the attention mechanism as a bias, enhancing the GT's capacity to model graph-structured data. 
Notably, applying PEs, which are typically concatenated with input features, would affect attention scores and could be considered as a way to introduce attention bias, as illustrated in Figure~\ref{fig:model}. However, concatenated PEs and attention bias fail to directly and effectively guide attention scores to fully capture structural information in graphs. From the Figure~\ref{fig:attn_vis}, it is evident that the attention patterns in GTs augmented with PE or attention bias exhibit minimal deviation from those observed in the vanilla GTs. This observation suggests that the inductive biases introduced by these methods have a limited effect on the model's attention mechanism. Furthermore, the attention in these models appears to be inattentive (characterized by dense patterns), potentially leading to a failure in focusing on key information and aggregating redundant information. These findings highlight a critical need for more sophisticated approaches capable of effectively incorporating structural insights into the self-attention framework of GTs. Therefore, the goal is to develop a method that not only enhances the model's attention mechanism by focusing on structurally significant features but also reduces redundancy in the information aggregation process.

Different from the aforementioned methods, we propose Gradformer, a novel method that innovatively integrates GTs with inductive bias. Specifically, Gradformer integrates an exponential decay mask into the GT self-attention architecture. This mask, multiplied with the attention score, explicitly controls each node's attention weights relative to other nodes, ensuring that the attention weights decay with an increasing node distance. In addition, the exponential decay ensures a gradual reduction in attention weights at the boundary of the full attention zone, avoiding an abrupt truncation. As a result, the introduced decay mask, rooted in the graph's structural bias, effectively guides the learning process within the self-attention framework. Furthermore, Gradformer incorporates a learnable constraint within the decay mask, applied to the attention heads, dynamically adjusting the node distance for full attention in a graph. This constraint amplifies the model's ability to discern local structural nuances and diversifies the attention heads, facilitating more effective assimilation of diverse structural information.

The design elements of Gradformer offer multiple benefits: \textbf{1)} The decay mask effectively integrates a form of prior knowledge, deeply connected to the graph's structural attributes, into the GT models. \textbf{2)} The structure-oriented prior knowledge precisely governs each node's interaction radius and hence defines the extent of its attention relative to other nodes, thus preventing the aggregation of redundant information. This is achieved by the design of exponential decay, which enables mask values for distant nodes approaching near-zero levels, thus effectively reducing their influence in the aggregation process. \textbf{3)} The exponential decay mask endows the enhanced attention mechanism in Gradformer with the ability to function as a unified form of GNNs and GTs, synergizing their strengths: the local processing power of GNNs and global aggregation capabilities of GTs (further expounded in Section~\ref{sec:discussion}).  In summary, Gradformer empowers the self-attention mechanism to effectively concentrate on structural information within the graph and limit unnecessary aggregation from distant nodes.

To evaluate the effectiveness of our Gradformer, we conduct extensive experiments on various commonly-used datasets, including the large-scale Open Graph Benchmark (OGB)~\cite{ogb-dataset}. The experimental results consistently demonstrate that Gradformer outperforms state-of-the-art GT models on most datasets with remarkable stability and accuracy improvements, even as network depth increases. This finding stands in stark contrast to the performance of other GT models, where accuracy tends to decline with increased depth~\cite{deepgraph}.  Additionally, incorporating the decay mask design into Gradformer provides notable advantages for the graph classification task  in low-resource settings, underscoring the versatility and practicality of our model. Our main contributions are summarized as follows:

\begin{enumerate}[leftmargin=12pt]
  \item We propose a general GT with an exponential decay mask attention mechanism, termed Gradformer, which enhances the self-attention mechanism by fully leveraging the graph's structural information. Therefore, Gradformer maintains the ability to capture long-range information while prioritizing the local information of the graph, guided by an intense inductive bias.

  \item We conduct extensive experiments to compare Gradformer with 14 GNN and GT baseline models for the graph classification task on various real-world graph datasets, including OGB. The experimental results consistently validate the effectiveness of the proposed Gradformer.
\end{enumerate}

\section{Related Work}
\label{sec:related work}

\paragraph{Graph Transformers.} 
In recent years, many transformer variants have been applied to graph modeling. Unlike GNNs, transformers display competitive or even superior performance across various tasks. Dwivedi \textit{et al.}~\shortcite{graph-transformer} were the first to extend the transformer architecture to graphs and propose PEs~\cite{position-encoding}. Subsequently, numerous variants of GT have been proposed, making significant progress in graph-level tasks~\cite{grover,compt,graphtrans,sat,lgi-gt}. However, these methods are primarily designed for graph-level tasks due to the time and memory constraints of the self-attention layer, which requires $\mathcal{O}(n^2)$ complexity. Therefore, multiple studies~\cite{gophormer,mask-transformer,det,deformable-transformer,nodeformer,difformer,gapformer,gtsp} have been conducted to enhance GTs' scalability and efficiency, facilitating their application in node-level tasks. Additionally, GTs have been applied in various fields, including natural language processing~\cite{transformer-entity-alignment}, computer vision~\cite{transformer-3d,gtgan}, recommendation systems~\cite{transformer-recommendation,gformer}, multimodal contexts~\cite{m-dgt,multimodal-gt}, and reinforcement learning~\cite{gdt}. For a more detailed introduction, please refer to the recent GT reviews~\cite{graphgps,transformer-review}.

\paragraph{Prior Knowledge in Graph Transformer.}
Numerous efforts have been undertaken to integrate prior knowledge into GTs to augment their performance. This integration can be classified into three categories. \textit{1) Proposed position encodings:} SAN~\cite{san} suggests utilizing learned Laplace eigenfunctions as a substitute for the original PE.  GraphGPS~\cite{graphgps} summarizes possible positional and structural encodings for GTs. \textit{2) Improved self-attention:} Graphormer~\cite{graphormer-v1} suggests adding structural encodings to the attention metric as a bias to enhance GTs' structured-data modeling. Moreover, RT~\cite{rt} and EGT~\cite{egt} consider incorporating the edge vectors in their self-attention calculations. \textit{3) Modified architecture:} GraphTrans~\cite{graphtrans} introduces a transformer sub-network above a GNN layer. In GraphGPS, the transformer and GNN layers are placed in parallel. In this manuscript, our proposed method lies on the line of improving the self-attention learning.


\section{Methodology}
\label{sec:method}

\subsection{Preliminaries}

\paragraph{Notations.}  A graph $\mathcal{G}$ can be represented by an adjacency matrix $\mathbf{A} \in \{0, 1\}^{ n \times n}$ and a node feature matrix $\mathbf{X} \in \mathbb{R}^{ n \times d}$, where $n$ is the number of nodes, $d$ is the node feature dimension, and  $\mathbf{A}[i, j]=1$ if an edge between nodes $v_{i}$ and $v_{j}$ exists, otherwise, $\mathbf{A}[i, j]=0$.

\paragraph{Graph Transformer.} GTs~\cite{transformer,graphormer-v1} consist of two essential parts: a multi-head self-attention (MHA) module and a feed-forward network (FFN). Given the node embedding matrix  $\mathbf{H}^{(l)} \in \mathbb{R}^{ n \times d^{(l)}}$ in a graph, a single attention head is computed as follows: 
\begin{equation}
\mathbf{H}^{(l+1)}=\operatorname{softmax}\left(\frac{\mathbf{Q}^{(l)} (\mathbf{K}^{(l)})^{\top}}{\sqrt{d^{(l)}}}\right) \mathbf{V}^{(l)}, 
\label{eq:sha}
\end{equation}
where $\mathbf{H}^{(l+1)} \in \mathbb{R}^{n \times d^{(l+1)}}$ is the output matrix,  $d^{(l+1)}$ is the output hidden dimension, and $\mathbf{Q}^{(l)} \in \mathbb{R}^{n \times d^{(l)}}$, $\mathbf{K}^{(l)} \in \mathbb{R}^{n \times d^{(l)}}$, and $\mathbf{V}^{(l)} \in \mathbb{R}^{n \times d^{(l)}}$ are the query, key, and value vectors, respectively, which are the projection results of $\mathbf{H}^{(l)} \in \mathbb{R}^{ n \times d^{(l)}}$:
\begin{equation}
\mathbf{Q}^{(l)}=\mathbf{H}^{(l)} \mathbf{W}^Q; \mathbf{K}^{(l)}=\mathbf{H}^{(l)} \mathbf{W}^K; \mathbf{V}^{(l)}=\mathbf{H}^{(l)} \mathbf{W}^V ,
\end{equation}
where $\mathbf{W}^Q \in \mathbb{R}^{d^{(l)} \times d^{(l+1)}}, \mathbf{W}^K \in \mathbb{R}^{d^{(l)} \times d^{(l+1)}}$, and $\mathbf{W}^V \in \mathbb{R}^{d^{(l)} \times d^{(l+1)}}$ are projection matrices. Note that the above single-head self-attention module can be generalized into a MHA via the concatenation operation. 


\subsection{Proposed Method: Gradformer}
\label{sec:propose-method}

This subsection introduces the Gradformer method. First, we describe the fundamental Gradformer architecture with a primary focus on its central module, the exponential decay mask attention mechanism. The mechanism comprises two components: exponential decay and learnable constraints. Subsequently, we explore the Gradformer method in detail, with a focus on its general form and computational complexity.

\subsubsection{Architecture}
\label{sec:pool-arch}
As depicted in Figure~\ref{fig:model}, Gradformer extends the original GT architecture by incorporating a \textit{decay mask}. The decay mask, derived from the graph's structural information, introduces a strong inductive bias into the self-attention learning.

\begin{figure}[!t] 
\begin{center}
\includegraphics[width=1.0\linewidth]{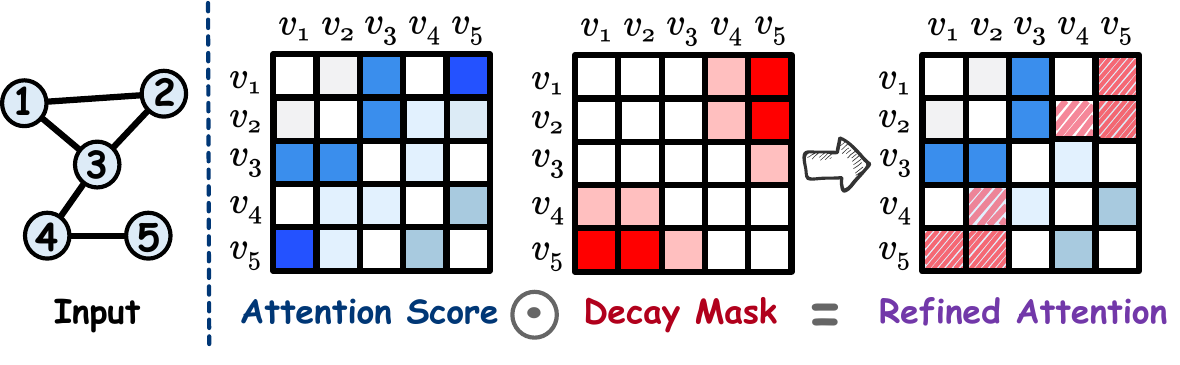}
\end{center}
\caption{The decay mask mechanism. The blue matrix represents the pairwise dot product, where the intensity of the blue color indicates the magnitude of the attention. The red matrix represents the decay mask, where the intensity of the red color indicates the magnitude of the mask. Once the mask is applied (indicated by red cells with diagonal stripes), the attention values in the masked cells become significantly attenuated. With this attention decay masking, the self-attention mechanism becomes more responsive to the graph's structural characteristics.}
\label{fig:model-2}
\end{figure}

\begin{figure}[!t] 
\begin{center}
\includegraphics[width=1.0\linewidth]{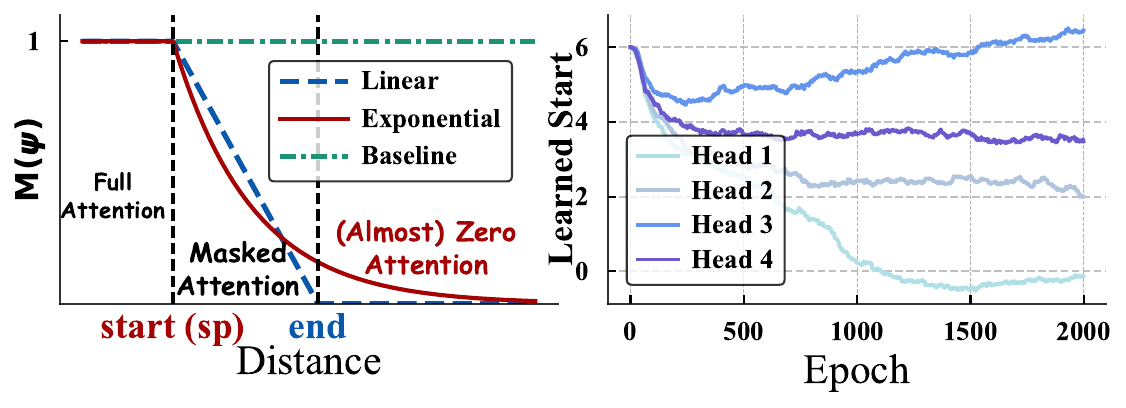}
\end{center}
\caption{\textbf{Left:} The value of the decay mask matrix $\mathbf{M}$ varies with the node-wise distance, $\psi$. Please note that the configuration of an {\color[HTML]{0A58AC} \textbf{end}} point is unique to linear decay, whereas exponential decay does not require such a parameter. Furthermore, the {\color[HTML]{A40203} \textbf{start}} point (\textit{i.e.}, $sp$ in Eq.~(\ref{eq:mask-3}))  is a learnable parameter, as depicted in the right part. \textbf{Right:} The learned start points for different attention heads demonstrate variation across epochs during training on the ZINC dataset.}
\label{fig:decay-line}
\end{figure}

\paragraph{Masking Attention with Exponential Decay.} 
According to Eq.~(\ref{eq:sha}), we observe that the attention matrix is comparable to a row-normalized adjacency matrix of a directly weighted complete graph. This dictates the aggregation of node features in a graph, similar to GCN. Unlike the input graph, this dynamically forms graph representations through the attention mechanism. However, the basic transformer does not directly consider the input structure (\textit{i.e.}, existing edges) when forming these weighted graphs (\textit{i.e.}, the attention matrices). Consequently, the attention mechanism may overlook the importance of certain neighbouring nodes due to feature similarity, and hence these dynamic graphs collapse immediately after aggregation. To address this issue, we introduce a decay mask $\mathbf{M}$, related to the graph's structural information, to participate in the aggregation process as follows: 
\begin{equation}
\mathbf{M} =\lambda^{\psi(v_i, v_j)} \in \mathbb{R}^{n \times n},
\label{eq:mask}
\end{equation}
where $\lambda$ is a scalar and $\psi\left(v_i, v_j\right)$ refers to a function that measures the spatial distance between $v_i$ and $v_j$  in a graph. The function $\psi(\cdot)$ can be defined by the connectivity between the nodes in the graph. In this paper, we choose $\psi\left(v_i, v_j\right)$ to present the distance of the shortest path (SPH) between $v_i$ and $v_j$ if the two nodes are connected. If not, we set the output of $\psi(\cdot)$ to a special value (\textit{i.e.}, -1). Based on the study~\cite{retentive-network}, our approach configures the attention mask $\mathbf{M}$ to decrease monotonically and exponentially, as depicted in the left part of Figure~\ref{fig:decay-line}. Compared to methods with linear decay, exponential decay eliminates the need for an additional endpoint and effectively filters attention at extended distances, with values for distant nodes diminishing to nearly zero. After obtaining the attention mask, it is multiplied with the attention score $\mathbf{S} \in \mathbb{R}^{n \times n}$, as illustrated in Figure~\ref{fig:model-2}, and formally defined as:
\begin{equation}
\begin{aligned}
\tilde{\mathbf{S}}^{(l)}=\operatorname{softmax}\left(\mathbf{S}^{(l)} \odot \mathbf{M}^{(l)}  \right), \\
\operatorname{Attn}\left(\mathbf{Q}^{(l)}, \mathbf{K}^{(l)}, \mathbf{V}^{(l)}\right)  =\tilde{\mathbf{S}}^{(l)} \mathbf{V}^{(l)}, 
\label{eq:mask-2}
\end{aligned}
\end{equation}
where $\tilde{\mathbf{S}}^{(l)}$ denotes the refined attention at layer $l$, and $\operatorname{Attn}$ denotes the self-attention operation with decay mask.

\paragraph{Mask Decay with Learnable Constraints.}  To enhance the attention mechanism's ability to model local structural information, given the strong locality inherent in graphs, a refined version of Eq.~(\ref{eq:mask}) is provided. This version incorporates a distance constraint into the decay operation, effectively tailoring the attention mechanism to be more sensitive to the local information. The refined decay mask is as follows:
\begin{equation}
\mathbf{M} =\lambda^{\operatorname{Relu}\left(\psi \left(v_i, v_j\right) - sp\right)},
\label{eq:mask-3}
\end{equation}
where $sp$ is a scalar that defines the starting point of the decay. Specifically, if the spatial distance in $\psi \left(v_i, v_j\right)$ is less than $sp$, the attention score between $v_i$ and $v_j$ remains unaffected by the decay. Hence, this design accurately controls each node's interaction radius, delineating the extent of its attention relative to other nodes and preventing the aggregation of redundant information.

Furthermore, to enable various attention heads to capture diverse structural information, we make the constraints $sp$ learnable. Consequently, when the decay mechanism is extended to multi-head attention, a decay mask bearing different constraints is applied to each head, as shown below:
\begin{equation*}
\tilde{\mathbf{S}}^{(l)}=\left[\mathbf{S}^{(l,1)} \odot \mathbf{M}^{(l,1)} ; \hat{\mathbf{S}}^{(l,2)} \odot \mathbf{M}^{(l,2)}; \cdots \hat{\mathbf{S}}^{(l,h)} \odot \mathbf{M}^{(l,h)}\right],
\end{equation*}
where $h$ represents the total number of attention heads. As illustrated in the right part of Figure~\ref{fig:decay-line}, the constraints of different attention heads are learned to capture various structural information. Specifically, we observe that the learned constraint for Head 3 is relatively large. This characteristic enables Head 3 to focus on global information of the graph. Conversely, Head 1 exhibits a different learning pattern, which allows model to capture proximate information.

\begin{table*}[!t]
\centering
\renewcommand\arraystretch{1.4} 
\setlength\tabcolsep{1.2pt} 
\resizebox{1.0\textwidth}{!}{%
\begin{tabular}{@{}l|ccccccccc@{}}
\toprule
\multirow{1}{*}{} & \multirow{1}{*}{\textbf{NCI1}} & \multirow{1}{*}{\textbf{PROTE.}} & \multirow{1}{*}{\textbf{MUTAG}} & \multirow{1}{*}{\textbf{COLLAB}} &  \multirow{1}{*}{\textbf{IMDB-B}} & \multirow{1}{*}{\textbf{PATTERN}} & \multirow{1}{*}{\textbf{CLUSTER}} & \multirow{1}{*}{\textbf{MOLHIV}}     & \multirow{1}{*}{\textbf{ZINC}}                    \\ \cmidrule(l){2-10}
\multirow{1}{*}{\textit{}} & \multirow{1}{*}{\textbf{Acc. $\uparrow$}} & \multirow{1}{*}{\textbf{Acc. $\uparrow$}} & \multirow{1}{*}{\textbf{Acc. $\uparrow$}} & \multirow{1}{*}{\textbf{Acc. $\uparrow$}} &  \multirow{1}{*}{\textbf{Acc. $\uparrow$}} & \multirow{1}{*}{\textbf{Acc. $\uparrow$}} & \multirow{1}{*}{\textbf{Acc. $\uparrow$}} & \multirow{1}{*}
{\textbf{AUROC. $\uparrow$}} & \multirow{1}{*}{\textbf{MAE $\downarrow$}}                        \\ \midrule
\multicolumn{9}{c}{\textit{GCN-based methods}}\\
\midrule
GCN~\cite{gcn}               & $79.68_{\pm 2.05}$    & $71.7_{\pm 4.7}$        & $73.4_{\pm 10.8}$      & $71.92_{\pm 1.18}$           & $74.3_{\pm 4.6}$     & $71.89_{\pm 0.33}$    & $69.50_{\pm 0.98}$  & $75.99_{\pm 1.19}$ & $0.367_{\pm 0.011}$      \\
GAT~\cite{gat}               & $79.88_{\pm 0.88}$    & $72.0_{\pm 3.3}$        & $73.9_{\pm 10.7}$      & $75.8_{\pm 1.6}$                 & $74.7_{\pm 4.7}$     & $78.27_{\pm 0.19}$    & $70.59_{\pm 0.45}$ & -- & $0.384_{\pm 0.007}$        \\
GIN~\cite{gin}            & $81.7_{\pm 1.7}$    & $73.76_{\pm 4.61}$        & $84.5_{\pm 8.9}$      & $73.32_{\pm 1.08}$                 & $75.1_{\pm 4.9}$     & $85.39_{\pm 0.14}$    & $64.72_{\pm 1.55}$  & $77.07_{\pm 1.49}$ & $0.526_{\pm 0.051}$      \\
GatedGCN~\cite{gatedgcn}            & $81.17_{\pm 0.79}$    & $74.65_{\pm 1.13}$       & $85.00_{\pm 2.67}$      & $80.70_{\pm 0.75}$                 & $73.20_{\pm 1.32}$     & $85.57_{\pm 0.09}$    & $73.84_{\pm 0.33}$  &-- &$0.282_{\pm 0.015}$      \\
\midrule
\multicolumn{9}{c}{ \textit{Graph Transformer-based methods}} \\
\midrule
GT~\cite{graph-transformer}               & $80.15_{\pm 2.04}$                                                 & $73.94_{\pm 3.78}$       & $83.9_{\pm 6.5}$             & $79.63_{\pm 1.02}$                    & $73.10_{\pm 2.11}$                             &$84.81_{\pm 0.07}$                       & $73.17_{\pm 0.62}$                   & -- & $0.226_{\pm 0.014}$                         \\
SAN~\cite{san}               & $80.50_{\pm 1.30}$                                                 & $74.11_{\pm 3.07}$      & $78.8_{\pm 2.9}$        & $79.42_{\pm 1.61}$     
& $72.10_{\pm 2.30}$                                         &$86.58_{\pm 0.04}$                       & $76.69_{\pm 0.65}$    &    $77.85_{\pm 2.47}$            &$0.139_{\pm 0.006}$                         \\
Graphormer~\cite{graphormer-v1}        & $81.44_{\pm 0.57}$                                              & $75.29_{\pm 3.10}$            &    $80.52_{\pm 5.79}$          &  ${\underline{{81.80}_{\pm {2.24}}}}$              &  $73.40_{\pm 2.80}$                               &     $86.65_{\pm 0.03}$                 &  $74.66_{\pm 0.24}$           &    $74.55_{\pm 1.06}$       & $0.122_{\pm 0.006}$                          \\ 
GraphTrans~\cite{graphtrans}        & $82.60_{\pm 1.20}$                                              & $75.18_{\pm 3.36}$            &    ${\underline{{87.22}_{\pm {7.05}}}}$          &  $79.81_{\pm 0.84}$          &  $74.50_{\pm 2.89}$                                  &   --                   &  --               &  $76.33_{\pm 1.11}$    & --                    \\ 
SAT~\cite{sat}               & $80.69_{\pm 1.55}$                                                 & $73.32_{\pm 2.36}$       & $80.50_{\pm 2.84}$           & $80.05_{\pm 0.55}$     & $75.90_{\pm 0.94}$                                            &$86.85_{\pm 0.04}$                       & $77.86_{\pm 0.10}$         & --           &$0.094_{\pm 0.008}$                       \\
EGT~\cite{egt}               & $81.91_{\pm 3.42}$                                                 & --       & --              &--                        & --                      &$86.82_{\pm 0.02}$                       & ${\textbf{79.23}_{\pm \textbf{0.35}}}$           &   ${\color[HTML]{808080}80.51^{\textbf{**}}_{\pm 0.30}}$         &$0.108_{\pm 0.009}$                       \\
GraphGPS~\cite{graphgps}        & ${\underline{{84.21}_{\pm {2.25}}}}$                                              & $\underline{{{75.77}_{\pm {2.19}}}}$            &    $85.00_{\pm 3.16}$        &  $81.40_{\pm 1.26}$             &  ${\textbf{77.40}_{\pm \textbf{0.63}}}$                                 &   $86.69_{\pm 0.06}$                    &  $78.02_{\pm 0.18}$          & $78.80_{\pm 0.49}$           & $0.070_{\pm 0.004}$                      \\ 
LGI-GT~\cite{lgi-gt}        & $82.18_{\pm 1.90}$                                              & --            &    --                  &  --              &  --                      &   ${\textbf{86.93}_{\pm \textbf{0.04}}}$                    &  $78.19_{\pm 0.10}$      & --            &  ${\textbf{0.069}_{\pm \textbf{0.002}}}$                        \\ 
KDLGT~\cite{kdlgt}        & --                                              & --            &    --                  &  --              &  --                      &   --                    &  --          & ${\underline{{78.98}_{\pm {1.78}}}}$         &  $0.130_{\pm 0.002}$                        \\ 
DeepGraph~\cite{deepgraph}        & --                                              & --            &    --                   &  --              &  --                    &   ${\color[HTML]{808080}90.66^{\textbf{*}}_{\pm 0.06}}$                    &  $77.91_{\pm 0.14}$      & --             & $0.072_{\pm 0.004}$                        \\ 
\midrule
\midrule
Gradformer (ours)       & ${\textbf{86.01}_{\pm \textbf{1.47}}} $   &  ${\textbf{77.50}_{\pm \textbf{1.86}}}$         &                         ${\textbf{88.00}_{\pm \textbf{2.45}}}$
   &  ${ \textbf{82.01}_{\pm \textbf{1.06}}}$
& ${\underline{{77.10}_{\pm 0.54}}}$    &   $ {\underline{{86.89}_{\pm {0.07}}}}$   &  $\underline{{78.55}_{\pm {0.16}}}$    &  ${\textbf{79.15}_{\pm \textbf{0.89}}}$         &  ${ \textbf{0.069}_{\pm \textbf{0.002}}}$                       \\ \bottomrule
\end{tabular}%
}
\begin{minipage}{1.0\linewidth} \small 
Notations: {\color[HTML]{808080} \textbf{*}:} DeepGraph utilizes a distinct evaluation metric on the pattern dataset, diverging from other models. Employing this metric, our method achieves an accuracy of $90.88_{\pm 0.06}$. {\color[HTML]{808080} \textbf{**}:} The result is derived from the fine-tuning of a large pretrained model.
\end{minipage} 
\caption{Experimental results on eight common datasets (the mean accuracy (\textbf{Acc.}), AUROC, and MAE, and standard deviation over 10 different runs). {\textbf{Bold}:} the best performance per dataset. {\underline{Underline}:} the second best performance per dataset.}
\label{tab:result-shallow}
\end{table*}


\subsection{Discussion} 
\label{sec:discussion}

\paragraph{Gradformer is a General Form of GNNs and GTs.} In essence, Gradformer represents a more generalized form of the GNN~\cite{gat} and GT~\cite{graph-transformer} models. This generalization becomes evident when considering the behavior of the decay mask $\mathbf{M}$ under varying settings of the parameter $\lambda$. Specifically, when $\lambda = 1$, the decay mask in Eq.~(\ref{eq:mask}) essentially transforms into an all-ones matrix. In this configuration, Gradformer effectively mirrors the GT model, as the decay mask exerts no modifying influence on the attention scores. Additionally, when $\lambda = 0$ , the decay mask $\mathbf{M}_{0}$ retains the mask value of the 1-hop neighbors as one and sets all others to zero. This configuration parallels the adjacency matrix $\mathbf{A}$, thereby aligning Gradformer closely with the GNN models. The formal derivation of the above process is as follows:
\begin{equation}
\mathbf{H}^{(l+1)}=\frac{1}{\sqrt{d^{(l)}}} \left(\mathbf{S}^{(l)} \odot \mathbf{M}_{0}^{(l)}\right) \mathbf{H}^{(l)}\mathbf{W} ; \quad \mathbf{M}_{0}^{(l)} = \mathbf{A} .
\label{eq:gradformer}
\end{equation}
Taking the GCN layer as an example, the message passing mechanism is expressed as follows:
\begin{equation}
\mathbf{H}^{(l+1)}  =\sigma\left(\mathbf{A} \mathbf{H}^{(l)} \mathbf{W}\right),
\label{eq:gnn}
\end{equation}
where $\sigma$ denotes the Sigmoid activation function. From these equations, we can deduce that Gradformer essentially constitutes a generalized GNN form, possessing at least the same expressive capability as traditional GNNs.

In summary, Gradformer surpasses traditional GNNs by broadening its receptive field to encompass more relevant nodes. Furthermore, compared to GTs, Gradformer demonstrates superior capacity in fusing node representations with graph structure, capturing more topological information.



\paragraph{Computational Complexity.} This subsection analyzes Gradformer's complexity. Compared to vanilla GTs, Gradformer incurs only the additional computational cost of point-wise multiplication between the decay and attention matrices. This process has significantly less computational complexity than self-attention. Additionally, Gradformer's additional memory requirement is the mask matrix, which shares the same size as the self-attention one. To illustrate Gradformer's complexity, detailed experimental results for time and memory are provided in the following section (See Figure~\ref{fig:efficiency}).




\section{Experiments}
\label{sec:experiment}


\subsection{Experimental Settings}

\paragraph{Datasets.} We utilize nine commonly-used real-world datasets from various sources to ensure diversity, including five graph datasets from the TU database~\cite{tu-dataset} (\textit{i.e.}, NCI1, PROTEINS, MUTAG, IMDB-B, and COLLAB), three datasets from Benchmarking GNN~\cite{benchmarkgnns} (\textit{i.e.}, PATTERN, CLUSTER, and ZINC), and one dataset from OGB~\cite{ogb-dataset} (\textit{i.e.}, OGBG-MOLHIV), involving diverse domains (\textit{e.g.}, synthetic, social, biology, and chemistry), sizes (\textit{e.g.}, ZINC and OGBG-MOLHIV are large datasets), and tasks (\textit{e.g.}, node classification, graph classification and regression). For all datasets, we strictly follow the evaluation metrics and dataset split recommended by the given benchmarks~\cite{graphormer-v1}. 


\paragraph{Baseline.} To demonstrate the effectiveness of our proposed method, we compare Gradformer with the following 14 baselines: \textbf{(\uppercase\expandafter{\romannumeral1}) 4 standard GCN-based models:} GCN~\shortcite{gcn}, GAT~\shortcite{gat}, GIN~\shortcite{gin}, and  GatedGCN~\shortcite{gatedgcn}; \textbf{(\uppercase\expandafter{\romannumeral2})  10 GT-based graph models:} GraphTransformer~\shortcite{graph-transformer}, SAN~\shortcite{san}, Graphormer~\shortcite{graphormer-v1}, GraphTrans~\shortcite{graphtrans}, SAT~\shortcite{sat},  EGT~\shortcite{egt}, GraphGPS~\shortcite{graphgps}, LGI-GT~\shortcite{lgi-gt}, KDLGT~\shortcite{kdlgt}, and DeepGraph~\shortcite{deepgraph}. For each baseline, we utilize the recommended settings as per the official implementation guidelines. 

\paragraph{Implementation Details.} We assess our proposed model's effectiveness by measuring its performance in graph classification and regression tasks. To ensure reliability, we conduct 10 trials for each model using different random seeds. Accordingly, we report the average test accuracy/AUROC/MAE based on the epoch when the best validation accuracy/AUROC/MAE is achieved.  Furthermore, all the experiments are conducted on a server equipped with 8 NVIDIA A100s. 



\begin{table}[!t]
\centering
\renewcommand\arraystretch{1.4} 
\setlength\tabcolsep{2.5pt} 
\resizebox{0.48\textwidth}{!}{%
\begin{tabular}{l|lccc}
\hline
Dataset  & \textbf{Method}    &\textbf{4 layers}             & \textbf{12 layers}            & \textbf{24 layers }                      \\ \hline
NCI1     & GraphGPS  & $84.21_{\pm 2.25}$                 & $84.09_{\pm1.68}$                 & $71.90_{\pm1.46}$                                   \\
         & \textbf{Ours}      & $\textbf{86.01}_{\pm 1.47}$                    & $\textbf{84.31}_{\pm 1.26}$                 & $\textbf{84.25}_{\pm 1.77}$                                \\ \hline\hline
  & \textbf{Method}    &\textbf{12 layers}             & \textbf{24 layers}            & \textbf{48 layers }                      \\ \hline
CLUSTER     & GraphGPS  & $\textbf{78.06}_{\pm 0.12}$                 & $\textbf{78.39}_{\pm 0.14}$                 & $70.91_{\pm 3.29}$                                 \\
         & \textbf{Ours}      & $77.78_{\pm 0.25}$                    & $78.31_{\pm 0.10}$                 & $\textbf{78.55}_{\pm 0.14}$                                 \\ \hline
PATTERN & GraphGPS  & \multicolumn{1}{l}{$86.74_{\pm 0.04}$} & \multicolumn{1}{l}{$86.68_{\pm 0.06}$} & \multicolumn{1}{l}{$85.83_{\pm 0.38}$}  \\
         & \textbf{Ours}      & \multicolumn{1}{l}{$\textbf{86.75}_{\pm 0.05}$} & \multicolumn{1}{l}{$\textbf{86.78}_{\pm 0.04}$} & \multicolumn{1}{l}{$\textbf{86.89}_{\pm 0.07}$} \\ \hline
\end{tabular}%
}
\caption{Performance of Gradformer with deep layers.}
\label{tab:result-deep}
\end{table}

\subsection{Overall Performance}
We evaluate the proposed model's effectiveness by comparing it with GT models that are both shallow and deep. For each model and dataset, we conduct 10 trials with random seeds, and then measure the mean accuracy and standard deviation, which are reported in Tables~\ref{tab:result-shallow} and~\ref{tab:result-deep}.

\paragraph{Performance of Gradformer.} From the results in Table~\ref{tab:result-shallow}, we observe that: 
\textbf{1)} Gradformer demonstrates exceptional performance, achieving state-of-the-art results on five datasets while remaining competitive on three others. This highlights the efficacy of the proposed method.
\textbf{2)} Notably, on small-scale datasets such as NCI1 and PROTEINS, Gradformer outperforms all 14 methods, exhibiting improvements of 2.13\% and 2.28\%, respectively. This demonstrates Gradformer's effective integration of inductive biases into the GT model, a crucial advantage especially in scenarios with limited data availability. To further validate this observation, additional experiments under low-resource conditions for the large-scale datasets have been conducted (see next section).
\textbf{3)} Furthermore, Gradformer delivers competitive results on large-scale datasets (\textit{e.g.}, ZINC), demonstrating its potential applicability across various dataset scales.

\begin{table}[!t]
\centering
\renewcommand\arraystretch{1.35} 
\setlength\tabcolsep{2pt} 
\resizebox{0.48\textwidth}{!}{%
\begin{tabular}{l|cccc}
\hline
NCI1 $\bm{\uparrow}$    & $\mathbf{5 \%}$ & $\mathbf{10 \%}$ & $\mathbf{2 5 \%}$ & $\mathbf{1 0 0 \%}$ \\ \hline
GraphGPS & $69.54_{\pm 0.96}$           & $74.70_{\pm 0.44}$         & $76.55_{\pm 1.19}$            & $84.21_{\pm 2.25}$               \\
\textbf{Ours}     & $\textbf{71.20}_{\pm 0.49}$           & $\textbf{76.38}_{\pm 0.85}$          & $\textbf{77.98}_{\pm 1.22}$             & $\textbf{86.01}_{\pm 1.47}$              \\ \hline \hline
ZINC $\bm{\downarrow}$ & $\mathbf{5 \%}$ & $\mathbf{10 \%}$ & $\mathbf{2 5 \%}$ & $\mathbf{1 0 0 \%}$ \\ \hline
GraphGPS & $0.438_{\pm 0.021}$          & $0.295_{\pm 0.012}$          & $0.182_{\pm 0.014}$          & $0.070_{\pm 0.004}$               \\
\textbf{Ours}     & $\textbf{0.429}_{\pm 0.019}$        & $\textbf{0.289}_{\pm 0.011}$         & $\textbf{0.171}_{\pm 0.009}$           & $\textbf{0.069}_{\pm 0.002}$         \\ \hline
\end{tabular}%
}
\caption{Results on Low-resource Settings. 5\% denotes the utilization of 5\% of the datasets as the training sets.}
\label{tab:result-few-shot}
\end{table}

\definecolor{lightpink}{HTML}{FAEAE1}
\definecolor{iceblue}{HTML}{E0F5FF}
\tcbset{
  pinkbox/.style={
    colback=lightpink,
    colframe=lightpink,
    width = 1.0cm,
    height = 0.35cm,
    halign=center, 
    valign=center, 
  }
}
\tcbset{
  bluebox/.style={
    colback=iceblue,
    colframe=iceblue,
    width = 1.0cm,
    height = 0.35cm,
    halign=center, 
    valign=center, 
  }
}
\begin{table}[!t]
\centering
\renewcommand\arraystretch{1.15} 
\setlength\tabcolsep{2pt} 
\resizebox{0.48\textwidth}{!}{%
\begin{tabular}{@{}l|cccc@{}}
\toprule
         & \textbf{NCI1}  & \textbf{PATTERN} & \textbf{CLUSTER} & \textbf{ZINC}  \\ \midrule
\textbf{GraphGPS} & $84.2$ & $86.68$    & $78.02$ & $0.070$ \\
\quad w/o MPNN  &  
$\textbf{80.04}_{\tiny \begin{tcolorbox}[pinkbox, left=0pt]{\text{$\downarrow \textbf{4.16}$}} \end{tcolorbox}} $   
& $71.01_{\tiny \begin{tcolorbox}[pinkbox, left=0pt]{\text{$\downarrow$ 15.67}} \end{tcolorbox}}$         
& $68.29_{\tiny \begin{tcolorbox}[pinkbox, left=0pt]{\text{$\downarrow$ 9.73}} \end{tcolorbox}}$ & $0.217_{\tiny \begin{tcolorbox}[pinkbox, left=0pt]{\text{$\downarrow$ 0.147}} \end{tcolorbox}}$     \\
\quad w/o PE    & $\textbf{83.67}_{\tiny \begin{tcolorbox}[bluebox, left=0pt]{\text{$\downarrow \textbf{0.53}$}} \end{tcolorbox}} $      & $86.63_{\tiny \begin{tcolorbox}[bluebox, left=0pt]{\text{$\downarrow$ 0.05}} \end{tcolorbox}} $         &$77.27_{\tiny \begin{tcolorbox}[bluebox, left=0pt]{\text{$\downarrow$ 0.75}} \end{tcolorbox}} $   & $0.113_{\tiny \begin{tcolorbox}[bluebox, left=0pt]{\text{$\downarrow$ 0.043}} \end{tcolorbox}} $    \\ \midrule \midrule
\textbf{Ours}     & $86.01$      & $86.56$         &$78.22$   &  $0.069$  \\
\quad w/o MPNN    & $80.80_{\tiny \begin{tcolorbox}[pinkbox, left=0pt]{\text{$\downarrow$ 5.21}} \end{tcolorbox}} $      &$\textbf{86.49}_{\tiny \begin{tcolorbox}[pinkbox, left=0pt]{\text{$\downarrow \textbf{0.07}$}} \end{tcolorbox}} $          & $\textbf{73.92}_{\tiny \begin{tcolorbox}[pinkbox, left=0pt]{\text{$\downarrow \textbf{4.30}$}} \end{tcolorbox}} $ & $\textbf{0.187}_{\tiny \begin{tcolorbox}[pinkbox, left=0pt]{\text{$\downarrow \textbf{0.118}$}} \end{tcolorbox}} $    \\
\quad w/o PE  & $85.01_{\tiny \begin{tcolorbox}[bluebox, left=0pt]{\text{$\downarrow$ 1.00}} \end{tcolorbox}}$ & $\textbf{86.70}_{\tiny \begin{tcolorbox}[bluebox, left=0pt]{\text{$\uparrow \textbf{0.14}$}} \end{tcolorbox}}$   & $\textbf{77.58}_{\tiny \begin{tcolorbox}[bluebox, left=0pt]{\text{$\downarrow \textbf{0.64}$}} \end{tcolorbox}}$& $0.116_{\tiny \begin{tcolorbox}[bluebox, left=0pt]{\text{$\downarrow$ 0.047}} \end{tcolorbox}}$\\ \bottomrule
\end{tabular}%
}
\caption{Analysis of Gradformer's performance without the MPNN or PE module. The downward arrow ($\downarrow$) denotes a reduction in model performance relative to the baseline method.}
\label{tab:result-ablation}
\end{table}

\paragraph{Performance of Gradformer with Deep Layers.} Table~\ref{tab:result-deep} displays the results of Gradformer with deep layers. Based on these results, the following observations can be made. \textbf{1)} Increasing the number of layers notably decreases the accuracy of deep GTs, such as GraphGPS (\textit{e.g.}, a drop of 14.4\% on NCI1 and 9.54\% on CLUSTER). \textbf{2)} In contrast, Gradformer demonstrates a significant improvement in this context.  It sustains performance on the NCI1 dataset and surpasses shallow models on CLUSTER and PATTERN datasets. This improvement can be attributed to Gradformer's inherent flexibility in accommodating increased model depth. Specifically, while self-attention in shallow networks is inclined to focus locally, it adopts a more global perspective in deeper networks. Therefore, incorporating residual connections enhances information flow between layers, stabilizing the learning process. The decay mask utilized by Gradformer plays a pivotal role in ensuring that the updated node information maintains higher fidelity to the original data, producing effects similar to those observed in shallower models.

\subsection{Further Discussions}
\label{sec:further-discussion}

\paragraph{Results on Low-resource Settings.}
We conduct experiments using datasets with a reduced number of labels on NCI1 and ZINC datasets. The results in Table~\ref{tab:result-few-shot} indicate that Gradformer consistently outperforms GraphGPS across all settings, particularly when the number of labels is very low, with improvements of 2.4\% and 2.3\% for 5\% and 10\% labeled data, respectively. This finding indicates that our method effectively incorporates the inductive biases into the GTs, enabling them to efficiently assimilate graph information with a limited number of labeled datasets. Furthermore, this discovery highlights the potential applicability of our model under resource-constrained conditions, such as the biomedical field where labeled data are scarce and costly.

\begin{figure}[!t] 
\begin{center}
\includegraphics[width=1.0\linewidth]{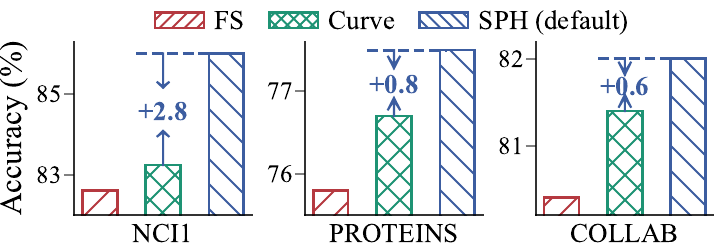}
\end{center}
\caption{Ablation study of graph structure index.}
\label{fig:ablation-sph}
\end{figure}

\begin{figure}[!t] 
\begin{center}
\includegraphics[width=1.0\linewidth]{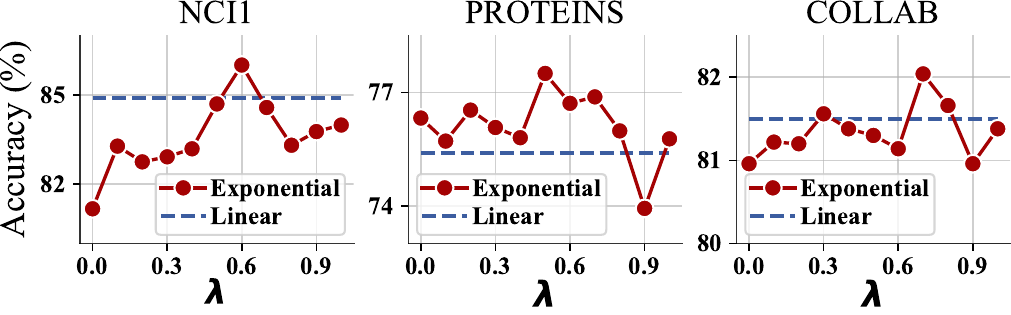}
\end{center}
\caption{The parameter analysis of the decay ratio, along with an ablation study of the decay function.}
\label{fig:para-decay-ratio}
\end{figure}

\paragraph{Performance without MPNN or Position Encoding.} In our subsequent analysis, we examine the impact of two key GT components: the Message Passing Neural Network (MPNN) module and PE. This investigation involves four graph datasets. It is important to note that except for the specific components under study, all other aspects of the models are maintained in line with the complete models. The findings, summarized in Table~\ref{tab:result-ablation}, indicate that: \textbf{1)} The removal of the MPNN module leads to a slight decline in Gradformer's performance, especially in contrast to GraphGPS. For instance, the performance of GraphGPS without MPNN diminishes by 15.67 on the PATTERN dataset, whereas Gradformer only experiences a reduction of 0.07. \textbf{2)} The absence of PE seems to have a minimal, and in some cases even beneficial, impact on Gradformer. The model demonstrates improved performance on certain datasets (\textit{i.e.}, PATTERN) when PE is omitted. This finding suggests that Gradformer's architecture, particularly its decay mask design, can  compensate for the absence of positional information. In summary, these results imply that our method effectively enables self-attention mechanisms to capture more structural information.

\paragraph{The Impact of Graph Structure Index.} As detailed in Section~\ref{sec:propose-method}, our model's decay mask is formulated based on specific graph structural indices. In addition to the previously mentioned SPH, our study also examines another structural index: the discrete Ricci curvature (Curve)~\cite{curve}. This index is the graph distance based on Riemannian manifold. Furthermore, we evaluate the effectiveness of a feature-based index, specifically the feature cosine similarity (FS). The results, illustrated in Figure~\ref{fig:ablation-sph}, show that SPH consistently outperforms the other two indexes. Notably, the accuracy achieved with FS is significantly lower than that obtained with the structural indices (\textit{i.e.}, SPH and Curve). These findings further emphasize the importance of incorporating structural information in enhancing model performance.

\paragraph{The Impact of Decay Ratio and Decay Function.} In an analysis conducted on three datasets, NCI1, PROTEINS, and COLLAB, the impact of varying the parameter $\lambda$ is examined. The results are depicted in Figure~\ref{fig:para-decay-ratio}. These findings reveal that as $\lambda$ increases, the model's accuracy initially rises, reaching optimal performance typically around the values of 0.5 or 0.6. Subsequently, the model's accuracy declines. Furthermore, as elaborated in Section~\ref{sec:discussion}, specific $\lambda$  values confer distinct characteristics to Gradformer. For instance, at $\lambda = 1$, Gradformer aligns with the conventional GT model, whereas at $\lambda = 0$,  it nearly regresses to a GNN model. Notably, across varying decay ratios, Gradformer consistently exhibits superior performance over the standard GNN model. In addition, observations indicate that the exponential decay function consistently outperforms the linear one.

\begin{figure}[!t] 
\begin{center}
\includegraphics[width=1.0\linewidth]{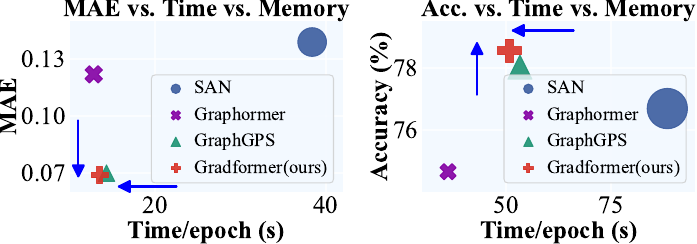}
\end{center}
\caption{Efficiency analysis comparing Gradformer with three baseline models on two datasets: ZINC (\textit{left}) and CLUSTER (\textit{right}). The size of the markers indicates GPU memory usage.  }
\label{fig:efficiency}
\end{figure}

\paragraph{Efficiency Analysis.} 
To validate Gradformer's efficiency, its training cost is compared with prominent methods such as SAN~\cite{san}, Graphormer~\cite{graphormer-v1}, and GraphGPS~\cite{graphgps}, with a focus on metrics such as running time and GPU memory usage. The comparative results are presented in Figure~\ref{fig:efficiency}. These findings reveal that Gradformer achieves an optimal balance between efficiency and effectiveness. Notably, Gradformer outperforms SAN and GraphGPS in computational efficiency and accuracy. Although Gradformer exhibits a marginally longer runtime compared to Graphormer, the former significantly surpasses the latter in accuracy, highlighting its superiority in balancing resource usage with high-performance outcomes.



\section{Conclusion}
In this study, we introduce Gradformer, a novel integration of GT with intrinsic inductive biases, achieved by applying an exponential decay mask with learnable parameters to the attention matrix. Through extensive experimentation across 9 graph datasets, Gradformer has shown its superiority by outperforming 14 contemporary GTs and GNNs. Notably, Gradformer's strength lies in its ability to maintain or even surpass shallow models in accuracy while deepening the network architecture. This is a feat not commonly observed in other GTs where accuracy tends to decline significantly in the same context. Despite its competitive performance, Gradformer still has areas for further improvement, including \textbf{1)} exploring the feasibility of achieving state-of-the-art structure without the use of MPNN, and \textbf{2)} investigating the potential for the decay mask operation to significantly improve GT efficiency.


\section*{Acknowledgments}
The work of Wenbin Hu was supported by the National Key Research and Development Program of China (2023YFC2705700). This work was supported in part by the Natural Science Foundation of China (No. 82174230), Artificial Intelligence Innovation Project of Wuhan Science and Technology Bureau (No. 2022010702040070), Natural Science Foundation of Shenzhen City (No. JCYJ20230807090211021).

\bibliographystyle{named_short}
\bibliography{reference_short}

\begin{thebibliography}{}

\bibitem[\protect\citeauthoryear{Cai \bgroup \em et al.\egroup }{2022}]{transformer-entity-alignment}
Weishan Cai, Wenjun Ma, Jieyu Zhan, and Yuncheng Jiang.
\newblock Entity alignment with reliable path reasoning and relation-aware heterogeneous graph transformer.
\newblock In {\em IJCAI}, 2022.

\bibitem[\protect\citeauthoryear{Chen and Li}{2022}]{m-dgt}
Sijia Chen and Baochun Li.
\newblock Multi-modal dynamic graph transformer for visual grounding.
\newblock In {\em CVPR}, 2022.

\bibitem[\protect\citeauthoryear{Chen \bgroup \em et al.\egroup }{2021}]{compt}
Jianwen Chen, Shuangjia Zheng, Ying Song, Jiahua Rao, and Yuedong Yang.
\newblock Learning attributed graph representation with communicative message passing transformer.
\newblock In {\em IJCAI}, 2021.

\bibitem[\protect\citeauthoryear{Chen \bgroup \em et al.\egroup }{2022}]{sat}
Dexiong Chen, Leslie O'Bray, and Karsten Borgwardt.
\newblock Structure-aware transformer for graph representation learning.
\newblock In {\em ICML}, 2022.

\bibitem[\protect\citeauthoryear{Choromanski \bgroup \em et al.\egroup }{2022}]{mask-transformer}
Krzysztof Choromanski, Han Lin, Haoxian Chen, Tianyi Zhang, Arijit Sehanobish, Valerii Likhosherstov, Jack Parker-Holder, Tamas Sarlos, Adrian Weller, and Thomas Weingarten.
\newblock From block-toeplitz matrices to differential equations on graphs: towards a general theory for scalable masked transformers.
\newblock In {\em ICML}, 2022.

\bibitem[\protect\citeauthoryear{Diao and Loynd}{2023}]{rt}
Cameron Diao and Ricky Loynd.
\newblock Relational attention: Generalizing transformers for graph-structured tasks.
\newblock In {\em ICLR}, 2023.

\bibitem[\protect\citeauthoryear{Ding \bgroup \em et al.\egroup }{2020}]{position-encoding}
Liang Ding, Longyue Wang, and Dacheng Tao.
\newblock Self-attention with cross-lingual position representation.
\newblock In {\em ACL}, 2020.

\bibitem[\protect\citeauthoryear{Dwivedi and Bresson}{2021}]{graph-transformer}
Vijay~Prakash Dwivedi and Xavier Bresson.
\newblock A generalization of transformer networks to graphs.
\newblock In {\em AAAI Workshop}, 2021.

\bibitem[\protect\citeauthoryear{Dwivedi \bgroup \em et al.\egroup }{2023}]{benchmarkgnns}
Vijay~Prakash Dwivedi, Chaitanya~K Joshi, Anh~Tuan Luu, Thomas Laurent, Yoshua Bengio, and Xavier Bresson.
\newblock Benchmarking graph neural networks.
\newblock {\em Journal of Machine Learning Research}, 2023.

\bibitem[\protect\citeauthoryear{Fey and Lenssen}{2019}]{pytorch-geometric}
Matthias Fey and Jan~E. Lenssen.
\newblock Fast graph representation learning with {PyTorch Geometric}.
\newblock In {\em ICLR Workshop}, 2019.

\bibitem[\protect\citeauthoryear{Guo \bgroup \em et al.\egroup }{2022}]{det}
Lingbing Guo, Qiang Zhang, and Huajun Chen.
\newblock Unleashing the power of transformer for graphs.
\newblock {\em arXiv:2202.10581}, 2022.

\bibitem[\protect\citeauthoryear{He and Wang}{2023}]{multimodal-gt}
Xuehai He and Xin Wang.
\newblock Multimodal graph transformer for multimodal question answering.
\newblock In {\em ACL}, 2023.

\bibitem[\protect\citeauthoryear{Hu \bgroup \em et al.\egroup }{2020}]{ogb-dataset}
Weihua Hu, Matthias Fey, Marinka Zitnik, Yuxiao Dong, Hongyu Ren, Bowen Liu, Michele Catasta, and Jure Leskovec.
\newblock Open graph benchmark: Datasets for machine learning on graphs.
\newblock {\em arXiv:2005.00687}, 2020.

\bibitem[\protect\citeauthoryear{Hu \bgroup \em et al.\egroup }{2023}]{gdt}
Shengchao Hu, Li~Shen, Ya~Zhang, and Dacheng Tao.
\newblock Graph decision transformer.
\newblock {\em arXiv preprint arXiv:2303.03747}, 2023.

\bibitem[\protect\citeauthoryear{Hussain \bgroup \em et al.\egroup }{2022}]{egt}
Md~Shamim Hussain, Mohammed~J. Zaki, and Dharmashankar Subramanian.
\newblock Global self-attention as a replacement for graph convolution.
\newblock In {\em SIGKDD}, 2022.

\bibitem[\protect\citeauthoryear{Kingma and Ba}{2014}]{adam}
Diederik~P Kingma and Jimmy Ba.
\newblock Adam: A method for stochastic optimization.
\newblock {\em arXiv:1412.6980}, 2014.

\bibitem[\protect\citeauthoryear{Kipf and Welling}{2017}]{gcn}
Thomas~N. Kipf and Max Welling.
\newblock Semi-supervised classification with graph convolutional networks.
\newblock In {\em ICLR}, 2017.

\bibitem[\protect\citeauthoryear{Kreuzer \bgroup \em et al.\egroup }{2021}]{san}
Devin Kreuzer, Dominique Beaini, William~L. Hamilton, Vincent L{\'e}tourneau, and Prudencio Tossou.
\newblock Rethinking graph transformers with spectral attention.
\newblock In {\em NeurIPS}, 2021.

\bibitem[\protect\citeauthoryear{Lai \bgroup \em et al.\egroup }{2023}]{curve}
Xin Lai, Yang Liu, Rui Qian, Yong Lin, and Qiwei Ye.
\newblock Deeper exploiting graph structure information by discrete ricci curvature in a graph transformer.
\newblock {\em Entropy}, 2023.

\bibitem[\protect\citeauthoryear{Li \bgroup \em et al.\egroup }{2016}]{gatedgcn}
Yujia Li, Richard Zemel, Marc Brockschmidt, and Daniel Tarlow.
\newblock Gated graph sequence neural networks.
\newblock In {\em ICLR}, 2016.

\bibitem[\protect\citeauthoryear{Li \bgroup \em et al.\egroup }{2023}]{gformer}
Chaoliu Li, Lianghao Xia, Xubin Ren, Yaowen Ye, Yong Xu, and Chao Huang.
\newblock Graph transformer for recommendation.
\newblock In {\em SIGIR}, 2023.

\bibitem[\protect\citeauthoryear{Liu \bgroup \em et al.\egroup }{2023}]{gapformer}
Chuang Liu, Yibing Zhan, Xueqi Ma, Liang Ding, Dapeng Tao, Jia Wu, and Wenbin Hu.
\newblock Gapformer: Graph transformer with graph pooling for node classification.
\newblock In {\em IJCAI}, 2023.

\bibitem[\protect\citeauthoryear{Liu \bgroup \em et al.\egroup }{2024}]{gtsp}
Chuang Liu, Yibing Zhan, Xueqi Ma, Liang Ding, Dapeng Tao, Jia Wu, Wenbin Hu, and Bo~Du.
\newblock Exploring sparsity in graph transformers.
\newblock {\em Neural Networks}, 2024.

\bibitem[\protect\citeauthoryear{Min \bgroup \em et al.\egroup }{2022}]{transformer-review}
Erxue Min, Runfa Chen, Yatao Bian, Tingyang Xu, Kangfei Zhao, Wenbing Huang, Peilin Zhao, Junzhou Huang, Sophia Ananiadou, and Yu~Rong.
\newblock Transformer for graphs: An overview from architecture perspective.
\newblock {\em arXiv:2202.08455}, 2022.

\bibitem[\protect\citeauthoryear{Morris \bgroup \em et al.\egroup }{2020}]{tu-dataset}
Christopher Morris, Nils~M Kriege, Franka Bause, Kristian Kersting, Petra Mutzel, and Marion Neumann.
\newblock Tudataset: A collection of benchmark datasets for learning with graphs.
\newblock {\em arXiv:2007.08663}, 2020.

\bibitem[\protect\citeauthoryear{Park \bgroup \em et al.\egroup }{2022}]{deformable-transformer}
Jinyoung Park, Seongjun Yun, Hyeonjin Park, Jaewoo Kang, Jisu Jeong, Kyung-Min Kim, Jung-woo Ha, and Hyunwoo~J Kim.
\newblock Deformable graph transformer.
\newblock {\em arXiv:2206.14337}, 2022.

\bibitem[\protect\citeauthoryear{Rampasek \bgroup \em et al.\egroup }{2022}]{graphgps}
Ladislav Rampasek, Mikhail Galkin, Vijay~Prakash Dwivedi, Anh~Tuan Luu, Guy Wolf, and Dominique Beaini.
\newblock Recipe for a general, powerful, scalable graph transformer.
\newblock In {\em NeurIPS}, 2022.

\bibitem[\protect\citeauthoryear{Rong \bgroup \em et al.\egroup }{2020}]{grover}
Yu~Rong, Yatao Bian, Tingyang Xu, Weiyang Xie, Ying WEI, Wenbing Huang, and Junzhou Huang.
\newblock Self-supervised graph transformer on large-scale molecular data.
\newblock In {\em NeurIPS}, 2020.

\bibitem[\protect\citeauthoryear{Sun \bgroup \em et al.\egroup }{2023}]{retentive-network}
Yutao Sun, Li~Dong, Shaohan Huang, Shuming Ma, Yuqing Xia, Jilong Xue, Jianyong Wang, and Furu Wei.
\newblock Retentive network: A successor to transformer for large language models.
\newblock {\em arXiv preprint arXiv:2307.08621}, 2023.

\bibitem[\protect\citeauthoryear{Tang \bgroup \em et al.\egroup }{2023}]{gtgan}
Hao Tang, Zhenyu Zhang, Humphrey Shi, Bo~Li, Ling Shao, Nicu Sebe, Radu Timofte, and Luc Van~Gool.
\newblock Graph transformer gans for graph-constrained house generation.
\newblock In {\em CVPR}, 2023.

\bibitem[\protect\citeauthoryear{Vaswani \bgroup \em et al.\egroup }{2017}]{transformer}
Ashish Vaswani, Noam Shazeer, Niki Parmar, Jakob Uszkoreit, Llion Jones, Aidan~N Gomez, \L~ukasz Kaiser, and Illia Polosukhin.
\newblock Attention is all you need.
\newblock In {\em NeurIPS}, 2017.

\bibitem[\protect\citeauthoryear{Veličković \bgroup \em et al.\egroup }{2018}]{gat}
Petar Veličković, Guillem Cucurull, Arantxa Casanova, Adriana Romero, Pietro Liò, and Yoshua Bengio.
\newblock Graph attention networks.
\newblock In {\em ICLR}, 2018.

\bibitem[\protect\citeauthoryear{Wu \bgroup \em et al.\egroup }{2021}]{graphtrans}
Zhanghao Wu, Paras Jain, Matthew Wright, Azalia Mirhoseini, Joseph~E Gonzalez, and Ion Stoica.
\newblock Representing long-range context for graph neural networks with global attention.
\newblock In {\em NeurIPS}, 2021.

\bibitem[\protect\citeauthoryear{Wu \bgroup \em et al.\egroup }{2022}]{nodeformer}
Qitian Wu, Wentao Zhao, Zenan Li, David Wipf, and Junchi Yan.
\newblock Nodeformer: A scalable graph structure learning transformer for node classification.
\newblock In {\em NeurIPS}, 2022.

\bibitem[\protect\citeauthoryear{Wu \bgroup \em et al.\egroup }{2023a}]{difformer}
Qitian Wu, Chenxiao Yang, Wentao Zhao, Yixuan He, David Wipf, and Junchi Yan.
\newblock {DIFF}ormer: Scalable (graph) transformers induced by energy constrained diffusion.
\newblock In {\em ICLR}, 2023.

\bibitem[\protect\citeauthoryear{Wu \bgroup \em et al.\egroup }{2023b}]{kdlgt}
Yi~Wu, Yanyang Xu, Wenhao Zhu, Guojie Song, Zhouchen Lin, Liang Wang, and Shaoguo Liu.
\newblock Kdlgt: A linear graph transformer framework via kernel decomposition approach.
\newblock In {\em IJCAI}, 2023.

\bibitem[\protect\citeauthoryear{Xu \bgroup \em et al.\egroup }{2019a}]{transformer-recommendation}
Chengfeng Xu, Pengpeng Zhao, Yanchi Liu, Victor~S. Sheng, Jiajie Xu, Fuzhen Zhuang, Junhua Fang, and Xiaofang Zhou.
\newblock Graph contextualized self-attention network for session-based recommendation.
\newblock In {\em IJCAI}, 2019.

\bibitem[\protect\citeauthoryear{Xu \bgroup \em et al.\egroup }{2019b}]{gin}
Keyulu Xu, Weihua Hu, Jure Leskovec, and Stefanie Jegelka.
\newblock How powerful are graph neural networks?
\newblock In {\em ICLR}, 2019.

\bibitem[\protect\citeauthoryear{Yin and Zhong}{2023}]{lgi-gt}
Shuo Yin and Guoqiang Zhong.
\newblock Lgi-gt: Graph transformers with local and global operators interleaving.
\newblock In {\em IJCAI}, 2023.

\bibitem[\protect\citeauthoryear{Ying \bgroup \em et al.\egroup }{2021}]{graphormer-v1}
Chengxuan Ying, Tianle Cai, Shengjie Luo, Shuxin Zheng, Guolin Ke, Di~He, Yanming Shen, and Tie-Yan Liu.
\newblock Do transformers really perform badly for graph representation?
\newblock In {\em NeurIPS}, 2021.

\bibitem[\protect\citeauthoryear{Zhao \bgroup \em et al.\egroup }{2021}]{gophormer}
Jianan Zhao, Chaozhuo Li, Qianlong Wen, Yiqi Wang, Yuming Liu, Hao Sun, Xing Xie, and Yanfang Ye.
\newblock Gophormer: Ego-graph transformer for node classification.
\newblock {\em arXiv:2110.13094}, 2021.

\bibitem[\protect\citeauthoryear{Zhao \bgroup \em et al.\egroup }{2023}]{deepgraph}
Haiteng Zhao, Shuming Ma, Dongdong Zhang, Zhi-Hong Deng, and Furu Wei.
\newblock Are more layers beneficial to graph transformers?
\newblock In {\em ICLR}, 2023.

\bibitem[\protect\citeauthoryear{Zhu \bgroup \em et al.\egroup }{2020}]{h2gcn}
Jiong Zhu, Yujun Yan, Lingxiao Zhao, Mark Heimann, Leman Akoglu, and Danai Koutra.
\newblock Beyond homophily in graph neural networks: Current limitations and effective designs.
\newblock In {\em NeurIPS}, 2020.

\bibitem[\protect\citeauthoryear{Zhu \bgroup \em et al.\egroup }{2021}]{transformer-3d}
Yiran Zhu, Xing Xu, Fumin Shen, Yanli Ji, Lianli Gao, and Heng~Tao Shen.
\newblock Posegtac: Graph transformer encoder-decoder with atrous convolution for 3d human pose estimation.
\newblock In {\em IJCAI}, 2021.

\end{thebibliography}

\clearpage
\appendix

\section{Experimental Settings}
\label{sec:implementaion}

We evaluate the effectiveness of the proposed model in terms of mean accuracy, mean absolute error and AUROC, conducting 10 trials with random seeds for each model. For each baseline, we refer to the recommended settings in the official implementations and implement Gradformer with Python (3.11.0), Pytorch (2.1.0), and Pytorch Geometric (2.4.0). All experiments are conducted on a Linux server with 8 NVIDIA A100s with 40GB memory.
\paragraph{Dataset Splits.} For TUDataset, we randomly split the dataset into training, validation, and test sets by a ratio of 8:1:1. The other evaluated benchmarks already define a standard train/validation/test dataset split. 

\paragraph{Training Details.} The Adam~\cite{adam} optimizer is used for the optimization of TUDataset without a learning rate scheduler. On the Benchmarking GNN and OGB datasets, we adopt AdamW optimizer with default settings, $\beta_1 = 0.9, \beta_2 = 0.999, \epsilon=10^{-8}$, along with linear warmup increase of learning rate at the beginning of training and cosine decay during training. Finally, we report the test metric (\textit{i.e.}, MAE, Accuracy, and AUROC) achieved by the epoch that gives the highest metric on the validation dataset.

\paragraph{Hyperparameters.} In our experimental setup for Gradformer, certain hyperparameters have been standardized to streamline the tuning process. Specifically, we have set the attention dropout rate at 0.5, and the weight decay is fixed at $1e^{-5}$. For other hyperparameters across various datasets, we have aligned our settings with those used in GraphGPS~\cite{graphgps} to maintain consistency and comparability. In the deep layer performance experiment, as detailed in Table 2 of the main text, the only variable we alter is the number of layers in Gradformer. This approach allows us to isolate the impact of layer depth on model performance. Additionally, we introduce a range of values for the hyperparameter $\lambda$, which is a key component in our model. The values for $\lambda$ that we explore are within the set $\{0.3, 0.4, 0.5, 0.6, 0.7\}$.  For a comprehensive understanding of our experimental setup, detailed hyperparameters are systematically listed in Table~\ref{tab:hyperpara}. In this table, the term `Layers' refers to the number of layers in Gradformer. `PE' denotes Positional Encoding, `RWSE' stands for Random-walk Structure Encoding, and `Lap' signifies Laplacian Eigenvectors Encodings. The notation `-20' alongside these encodings indicates that the size of the positional embedding is set to 20. 

\begin{table*}[!h]
    \setlength\tabcolsep{4pt} 
    \centering
    \footnotesize
    \renewcommand\arraystretch{1.4} 
    \setlength\tabcolsep{6pt} 
    \begin{tabular}{l|ccccccccc}\toprule
     & \textbf{NCI1} &\textbf{PROTEINS} &\textbf{MUTAG} &\textbf{COLLAB} &\textbf{IMDB-B} &\textbf{CLUSTER} &\textbf{PATTERN} &\textbf{MOLHIV} &\textbf{ZINC}\\\midrule
    Layers& 4  & 4  & 4 & 4  & 4 & 48 & 48 & 10 & 10\\
    Hidden dim& 64 & 64 & 64 & 64 & 64 & 48 & 64 & 64 & 64 \\
    Batch size& 128 & 128 & 128 & 128 & 128 & 16 & 32 & 32 & 32\\
    Dropout& 0 & 0 & 0 & 0 & 0 & 0.1 & 0 & 0.05 & 0  \\
    Heads& 4 & 4 & 4 & 4 & 4 & 8 & 4 & 4 & 4 \\
    PE & RWSE-20 & RWSE-20 & RWSE-20 & RWSE-20& RWSE-20& Lap-10 & None & RWSE-16&  RWSE-20 \\
    $\lambda$& [0.6, 0.7] & [0.5, 0.6] & 0.7 & 0.7 & 0.7 & 0.3 & 0.3 & 0.7 & 0.6 \\
    \bottomrule
    \end{tabular}
    \caption{Main experiment hyperparameters in the main text.}
    \label{tab:hyperpara}
\end{table*}

\section{Datasets and Baselines}
\label{sec:experiments-setting}

\subsection{Introduction of Datasets}

We use a total of nine datasets, including five from TUDataset~\cite{tu-dataset}, three from Benchmarking GNN~\cite{benchmarkgnns}, and one from Open Graph Benchmark (OGB)~\cite{ogb-dataset}, which vary in content domains, sizes, and tasks. All the adopted graph datasets can be downloaded from PyTorch Geometric (PyG). The dataset statistics are summarized in Table~\ref{tab:ds_summary}. A brief introduction to these datasets is provided below.

\begin{table*}[!t]
    \setlength\tabcolsep{4pt} 
    \centering
    \footnotesize
    \renewcommand\arraystretch{1.5} 
    \setlength\tabcolsep{6pt} 
    \begin{tabular}{c|rrrrcccc}\toprule
    \textbf{Source} & \textbf{Dataset} &\textbf{\# Graphs} &\textbf{\# Nodes} &\textbf{\# Edges} &\textbf{Predict level} &\textbf{Predict task} &\textbf{Metric} \\\midrule
    \multirow{5}{*}{\parbox{3cm}{\centering\textbf{TUDataset}\\\cite{tu-dataset}}}
    &NCI1 & 4,110 & 29.8 & 32.3 &graph &2-class classif. & Accuracy \\
    &PROTEINS & 1,113 & 39.1 & 72.8 &graph &2-class classif. & Accuracy \\
    &MUTAG & 188 & 17.9  & 19.7 &graph &2-class classif. & Accuracy \\
    &COLLAB & 5,000 & 74.5  & 2,457.7 &graph &3-class classif. & Accuracy \\
    &IMDB-B & 1,000 & 19.8  & 96.5 &graph &2-class classif. & Accuracy 
    \\\midrule
    \multirow{3}{*}{\parbox{3cm}{\centering\textbf{Bench. GNN}\\\cite{benchmarkgnns}}}
    &ZINC &12,000 &23.2 &24.9 &graph &regression &MAE \\
    &PATTERN &14,000 &118.9 &3,039.3 &inductive node &2-class classif. &Accuracy \\
    &CLUSTER &12,000 &117.2 &2,150.9 &inductive node &6-class classif. &Accuracy
    \\\midrule
    \textbf{OGB}~\cite{ogb-dataset}&ogbg-molhiv &41,127 &25.5 &27.5 &graph &2-class classif. &AUROC \\
    \bottomrule
    \end{tabular}
     \caption{Overview of the datasets mentioned in the main text.}
    \label{tab:ds_summary}
\end{table*}
\vspace{8pt}
\noindent \textbf{NCI1} consists of 4,110 molecule graphs from TUDataset, which represent two balanced subsets of datasets for chemical compounds screened for activity against non-small cell lung cancer and ovarian cancer cell lines, respectively.  \vspace{10pt} \\
\noindent \textbf{PROTEINS} consists of 1,113 protein graphs from TUDataset, where each graph corresponds to a protein molecule, nodes represent amino acids, and edges capture the interactions between amino acids. \vspace{10pt} \\
\noindent \textbf{MUTAG} consists of 188 chemical compounds from TUDataset, divided into two classes according to their mutagenic effect on a bacterium. \vspace{10pt} \\
\noindent \textbf{IMDB-BINARY} consists 1,000 movie graphs from TUDataset, classifying movies into two categories based on certain criteria. Each graph represents a movie, with nodes as actors and edges indicating their collaborations. \vspace{10pt} \\
\noindent \textbf{COLLAB} consists 5,000 graphs representing collaboration networks from TUDataset. Nodes represent authors, and edges denote collaborations between them. The task involves classifying graphs into three classes based on specific criteria. \vspace{10pt} \\
\noindent \textbf{ZINC} consists of 12,000 chemical compounds represented as graphs. The task associated with this dataset is regression, constraining solubility of the molecule. \vspace{10pt} \\
\noindent \textbf{CLUSTER} consists of 12,000 graphs. This dataset is widely employed to model communities in social networks by capturing intra- and extra-community relationships. It is generated with the Stochastic Block Model (SBM). \vspace{10pt} \\
\noindent \textbf{PATTERN} consists of 14,000 graphs and aims to identify nodes within a graph associated with one of 100 distinct sub-graph patterns. These patterns are randomly generated using SBM parameters different from the remainder of the graph. Different from other tasks, PATTERN and CLUSTER are node classification tasks. \vspace{10pt} \\
\noindent \textbf{Ogbg-molhiv} consists of 41,127 graphs, which are molecular property prediction datasets adopted by OGB from MoleculeNet. The prediction task involves 2-class classification of a molecule’s fitness to inhibit HIV replication.

\subsection{Introduction of Baselines}
To demonstrate the effectiveness of our proposed method, we compare Gradformer against 14 baselines. These baselines consist of four general Graph Convolutional Network (GCN)-based models and 10 Transformer-based models designed for graph-related tasks. We use the original paper implementations for all baseline models. If the original paper does not provide specific dataset settings, we maintain consistent parameter configurations with Gradformer. The links to the code implementations are included in Table~\ref{tab:baselines}; if unavailable, we utilize the implementations from PyTorch Geometric. 

\begin{table}[!t]
\centering
\renewcommand\arraystretch{1.4} 
\setlength\tabcolsep{2.5pt} 
\resizebox{0.48\textwidth}{!}{%
\begin{tabular}{@{}c|c@{}}
\toprule
\textbf{ Models }                & \textbf{ Code Links}                                                                                                 \\ \midrule
\multicolumn{2}{c}{ \textit{GCN-based methods}} \\
\midrule
GCN~\shortcite{gcn} & \url{https://github.com/tkipf/gcn} \\
GAT~\shortcite{gat} & \url{https://github.com/PetarV-/GAT} \\
GIN~\shortcite{gin} & \url{https://github.com/weihua916/powerful-gnns} \\
GatedGCN~\shortcite{gatedgcn} & \url{https://github.com/pyg-team/pytorch\_geometric} \\
\midrule\multicolumn{2}{c}{ \textit{Graph Transformer-based methods}} \\
\midrule
GT~\shortcite{graph-transformer} & \url{https://github.com/graphdeeplearning/graphtransformer}                               \\
SAN~\shortcite{san}                    & \url{https://github.com/DevinKreuzer/SAN}                                                 \\
Graphormer~\shortcite{graphormer-v1}             & \url{https://github.com/Microsoft/Graphormer}    
\\  
GraphTrans~\shortcite{graphtrans}             & \url{https://github.com/ucbrise/graphtrans}    
\\  
SAT~\shortcite{sat}             
& \url{https://github.com/BorgwardtLab/SAT/tree/main}    
\\  
EGT~\shortcite{egt}             
& \url{https://github.com/shamim-hussain/egt_pytorch}    
\\  
GraphGPS~\shortcite{graphgps} 
& \url{https://github.com/rampasek/GraphGPS/tree/main}    
\\  
LGI-GT~\shortcite{lgi-gt} 
& \url{https://github.com/shuoyinn/LGI-GT} 
\\
KDL-GT~\shortcite{kdlgt} 
& --   
\\
DeepGraph~\shortcite{deepgraph} 
& \url{https://github.com/zhao-ht/DeepGraph/tree/master}
\\
\bottomrule
\end{tabular}%
}
\begin{minipage}{1.0\linewidth} \small 
\vspace{2pt}
\end{minipage}
\caption{Baselines and their URLs.}
\label{tab:baselines}
\end{table}

\subsubsection{5.2.1 \quad GCN-based Methods} \vspace{5pt}

\noindent \textbf{GCN}~\cite{gcn}. Each node aggregates information from its neighboring nodes, facilitating the network in capturing the relational dependencies within the graph. This mechanism allows for learning complex patterns and representations for each node in the graph. \vspace{10pt} \\
\noindent \textbf{GAT}~\cite{gat}. It operates on graph-structured data and utilizes self-attention layers of masks. Such a utilization is computationally efficient in these networks, enabling the assignment of varying importance to different nodes when processing regions of different sizes.\vspace{10pt} \\
\noindent \textbf{GIN}~\cite{gin}. GIN is a streamlined yet potent network architecture that shows comparable graph structure discrimination to the Weisfeiler-Lehman test and exhibits robust representation capabilities. \vspace{10pt} \\
\textbf{GatedGCN}~\cite{gatedgcn}. It incorporates gating mechanisms to enhance the learning process within the graph structure by selectively filtering and weighting the information propagated through the edges of the graph. 
\subsubsection{5.2.2 \quad Graph Transformer-based Methods} \vspace{5pt} 

\textbf{GT}~\cite{graph-transformer}. GT is the pioneering work that generalizes transformer networks to arbitrary graphs and introduces the corresponding architecture. \vspace{10pt} \\ 
\textbf{SAN}~\cite{san}. SAN combines node features with learned position encodings (from Laplacian eigenvalues and eigenvectors), then uses them as attention keys and queries respectively, and employs edge features as attention Force values. \vspace{10pt} \\
\textbf{Graphormer}~\cite{graphormer-v1}. The Graphormer model introduces three structural encodings to enable the Transformer model to capture the structural information of the graph. These encodings allow the self-attention layer of the Graphormer model to successfully capture more important nodes or node pairs,leading to more accurate attention weight allocation. \vspace{10pt} \\
\textbf{GraphTrans}~\cite{graphtrans}. GraphTrans designed a novel GNNs readout module that uses a special token to aggregate all pairs interactions into a classification vector. It has been proven that modeling pairs node-node interaction is particularly important for large graph classification tasks. \vspace{10pt} \\
\textbf{SAT}~\cite{sat}. SAT proposes Structure-Aware Transformer, a simple and flexible graph Transformer based on a novel self-attention mechanism. This mechanism integrates structural information into the original self-attention by extracting a subgraph representation rooted at each node before computing attention. \vspace{10pt} \\
\textbf{EGT}~\cite{egt}. EGT exclusively use global self-attention as an aggregation mechanism rather than static localized convolutional aggregation.\vspace{10pt} \\
\textbf{GraphGPS}~\cite{graphgps}. GraphGPS allows for combining message passing networks with linear (long-range) transformer models to create a hybrid network. \vspace{10pt} \\
\textbf{LGI-GT}~\cite{lgi-gt}. LGI-GT introduces a novel method for global information representation by propagating [CLS] token embeddings. It also incorporates an effective message passing module called edge-enhanced local attention, making LGI-GT a fully attentive Graph Transformer. \vspace{10pt} \\
\textbf{KDL-GT}~\cite{kdlgt}. KDLGT utilizes kernel decomposition methods to rearrange the order of matrix multiplications, thereby reducing the complexity to linear levels.
\vspace{10pt} \\
\textbf{DeepGraph}~\cite{deepgraph}. DeepGraph is a novel graph transformer model that explicitly incorporates substructure tokens in the encoded representation. It applies local attention to related nodes, yielding substructure-based attention encoding.


\section{Evaluation on Other Task}
\label{sec:node_cls}


Given Gradformer's exemplary performance in graph classification task, we are motivated to extend our evaluation to node classification task. This shift in focus aims to specifically assess Gradformer's capability to leverage local information within a graph, a crucial aspect of node classification.  Traditional GTs typically treat all nodes in a graph uniformly, primarily relying on positional encoding to capture and preserve the graph's structural information. This approach, while effective in certain contexts, may not adequately emphasize the nuances of local graph structures. In contrast, the decay mask mechanism integrated into Gradformer offers significant advancements.

\subsection{Introduction of Datasets}

We use three graph datasets, including two homophilic datasets (\textit{i.e.}, Cora and Citeseer) and one heterophilic dataset (\textit{i.e.}, Actor). All mentioned datasets are available from PyTorch Geometric (PyG)~\cite{pytorch-geometric}. The detailed statistics for the node classification datasets are shown in Table~\ref{tab:node_dataset}. 

\begin{table}[!h]
    \setlength\tabcolsep{4pt} 
    \centering
    \footnotesize
    \renewcommand\arraystretch{1.5} 
    \setlength\tabcolsep{4.5pt} 
    \begin{tabular}{l|ccccc}\toprule
    \textbf{Dataset}  &\textbf{\# Nodes} &\textbf{\# Edges}  &\textbf{Classes} &\textbf{Hom. ratio} & \textbf{Metric}\\\midrule
    \textbf{Cora} & 2,708 & 5,429 & 7 & 0.81 & Accuracy\\
    \textbf{Citeseer} & 3,327 & 4,732 & 6 & 0.74 & Accuracy\\
    \textbf{Actor} & 7,600 & 30,019 & 5 & 0.22 & Accuracy \\
    \bottomrule
    \end{tabular}
    \caption{Overview of the node classification datasets.}
    \label{tab:node_dataset}
\end{table}
\noindent \textbf{Definition of Homophily and Heterophily.}
The edge homophily ratio ($\gamma$) introduced in H2GCN~\cite{h2gcn} quantifies the level of graph homophily. The edge homophily ratio is defined as follows:
\begin{equation}
\gamma=\frac{\left|\left\{(u, v):(u, v) \in \varepsilon \wedge y_u=y_v\right\}\right|}{|\varepsilon|},
\end{equation}
where $u$ and $v$ represent the nodes in the graph, $y_u$ and $y_v$ refer to the node labels, and $\varepsilon$ represents the set of the edges in the graph. The edge homophily ratio, $\gamma$ $\rightarrow 0$, indicates strong heterophily in the graph; $\gamma$ $\rightarrow 1$ suggests strong homophily. \vspace{5pt} \\
\noindent \textbf{Homogeneous Graph.} In the Cora dataset, nodes represent academic documents with citation relationships serving as connections between nodes. In the Citeseer dataset, nodes represent computer science papers, and edges indicate citation relationships. Each paper node is labeled with a predefined category, reflecting various topics in computer science.\vspace{5pt} \\
\noindent \textbf{Heterogeneous Graph.} In the Actor dataset, nodes are linked to information regarding their participation in diverse movies or TV shows, and the edges represent connections between actors.

\subsection{Implementation Details.} We evaluate the effectiveness of the proposed strategy in terms of node classification accuracy and conduct 10 trials with random seeds for each model. All methods share the same training, validation, and test splits, and the average and standard derivation are reported as the performance. The hyperparameter learning rate is set to 0.0001, with both the training batch size and evaluation batch size set to 128. Training epoch is set to 200. $\lambda$ is searched within $\{0.4, 0.5, 0.6, 0.7, 0.8\}$, and the layer of Gradformer is searched within $\{1, 2, 3, 4\}$.  Subsequently, we select GraphGPS~\cite{graphgps} as the baseline to evaluate the effectiveness of Gradformer. Finally, we report the test accuracy achieved at the epoch with the highest accuracy on the validation dataset.

\subsection{Experimental Results.} Table~\ref{tab:node_res} provides a comprehensive summary of the node classification results. Based on these results, we can draw the following conclusions:
\textbf{1) Superiority of Gradformer Over Baselines: } Gradformer demonstrates a consistent improvement in performance over the baseline models across three different datasets: Cora, Citeseer, and Actor. Specifically, the accuracy enhancements observed on these datasets are 0.12$\%$, 0.52$\%$, and 1.06$\%$, respectively. This consistent outperformance across diverse datasets highlights the effectiveness of the Gradformer model in node classification tasks. \textbf{2) Enhanced Performance on Heterogeneous Graphs:} A notable observation from the results is that Gradformer exhibits particularly impressive performance on the Actor dataset, which is a heterogeneous graph. The significantly better results on this dataset, compared to Cora and Citeseer, suggest that Gradformer is particularly well-suited for handling heterogeneous graphs. 

\begin{table}[!h]
    \setlength\tabcolsep{4pt} 
    \centering
    \footnotesize
    \renewcommand\arraystretch{1.5} 
    \setlength\tabcolsep{4.5pt} 
    \begin{tabular}{l|ccc}\toprule
     &\textbf{Cora} &\textbf{Citeseer}  &\textbf{Actor} \\\midrule
    GraphGPS & $81.57_{\pm 1.54}$ & $71.10_{\pm 2.88}$ & $32.57_{\pm 0.76}$ \\
    \textbf{Ours} & $\textbf{81.69}_{\pm \textbf{1.21}}$ &$\textbf{71.62}_{\pm \textbf{2.80}}$& $\textbf{33.63}_{\pm \textbf{0.75}}$\\
    \bottomrule
    \end{tabular}
    \caption{Result of the node classification.}
    \label{tab:node_res}
\end{table}


\section{The Pseudocode of Decay Mask} 
\label{sec:code}

\begin{figure}[!h] 
\begin{center}
\includegraphics[width=0.99\linewidth]{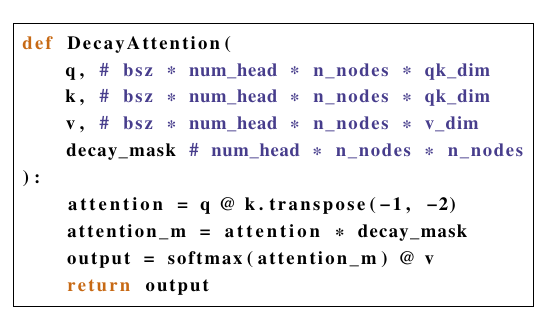}
\end{center}
\caption{The Pseudocode of decay mask in Gradformer.}
\label{fig:code}
\end{figure}

The most crucial aspect of implementing Gradformer is the addition of the decay mask to the attention scores. This process involves simply multiplying the computed attention scores with the preprocessed decay mask. The pseudocode for adding the decay mask is presented in Figure~\ref{fig:code}, highlighting the method's straightforwardness and simplicity and thus underscoring its implementation efficiency.

\end{document}